\documentclass[journal]{IEEEtran}
\ifCLASSINFOpdf
\else
\fi
\usepackage{mathtools}
\usepackage{cite}
\everymath{\displaystyle}
\usepackage{graphicx}
\usepackage{epstopdf}
\usepackage{float}
\usepackage{authblk}
\usepackage{caption}
\usepackage{subcaption}
\usepackage{amssymb}
\usepackage{pifont}% http://ctan.org/pkg/pifont
\newcommand{\cmark}{\ding{51}}%
\newcommand{\xmark}{\ding{55}}
\usepackage{algorithm}
\usepackage{algorithmic}
\usepackage{lipsum}
\usepackage{booktabs}
\DeclareCaptionLabelSeparator{periodspace}{.\quad}
\begin{document}
\title{Fuzzy c-Shape: A new algorithm for clustering finite time series waveforms}

\author{Fateme~Fahiman,~\IEEEmembership{Student Member,~IEEE,}
        James~C.~Bezdek,~\IEEEmembership{Life Fellow,~IEEE,}
        Sarah~M.~Erfani,
        Christopher~Leckie,
        and~Marimuthu~Palaniswami,~\IEEEmembership{Fellow,~IEEE}% <-this % stops a space
\thanks{F. Fahiman and M. Palaniswami are with the Department
of Electrical and Electronic Engineering, University of Melbourne,
Australia, (e-mail: ffahiman@student.unimelb.edu.au; palani@unimelb.edu.au).}% <-this % stops a space
\thanks{J. C. Bezdek, C. Leckie and S. M. Erfani are with the Department of Computing and Information Systems, The University of Melbourne, Australia, (e-mail: \{jbezdek, sarah.erfani, caleckie\}@unimelb.edu.au).}}% <-this %
\maketitle

\begin{abstract}
The existence of large volumes of time series data in many applications has motivated data miners to investigate specialized methods for mining time series data. Clustering is a popular data mining method due to its powerful exploratory nature and its usefulness as a preprocessing step for other data mining techniques. This article develops two novel clustering algorithms for time series data that are extensions of a crisp c-shapes algorithm. The two new algorithms are heuristic derivatives of fuzzy c-means (FCM). Fuzzy c-Shapes plus (FCS+) replaces the inner product norm in the FCM model with a shape-based distance function. Fuzzy c-Shapes double plus (FCS++) uses the shape-based distance, and also replaces the FCM cluster centers with shape-extracted prototypes. Numerical experiments on 48 real time series data sets show that the two new algorithms outperform state-of-the-art shape-based clustering algorithms in terms of accuracy and efficiency. Four external cluster validity indices (the Rand index, Adjusted Rand Index, Variation of Information, and Normalized Mutual Information) are used to match candidate partitions generated by each of the studied algorithms. All four indices agree that for these finite waveform data sets, FCS++ gives a small improvement over FCS+, and in turn, FCS+ is better than the original crisp c-shapes method. Finally, we apply two tests of statistical significance to the three algorithms. The Wilcoxon and Friedman statistics both rank the three algorithms in exactly the same way as the four cluster validity indices.
\end{abstract}

% \fi
\IEEEpeerreviewmaketitle

%----------------------------------------------------------------------------------------------------------------------------------------------
%                    INTRODUCTION
%-------------------------------------------------------------------------------------------------------------------------------------------
\section{Introduction}
\IEEEPARstart{I}{ncreasing}, large volumes of time series data are becoming available from a diverse range of sources, including smart phones, environmental sensors and biological devices. Techniques for analyzing time series data include dimensionality reduction  \cite{jadid15}, indexing \cite{kshape41},\cite{kshape44}, and segmentation \cite{jadid16}. In addition, techniques have been developed to extract the underlying shape or trends in the data, for tasks such as pattern discovery \cite{kshape43, kshape64, kshape53}, classification\cite{kshape37,kshape68}, rule discovery and summarization. An open challenge in time series analysis is how to cluster a set of time series according to the similarity of their underlying shape. The basic objective of this article is to generalize the fuzzy c-means (FCM) model \cite{bibi3} to accommodate clustering in finite time series data.
\\\indent In general, clustering is an analytic tool that can identify interesting patterns and correlation in the underlying data  \cite{kshape33}. For example in ECG signals the goal may be to cluster the ECG signals so that normal heart signals can be distinguished from abnormal signals based on the shape and phase of the time series signals. A key problem in finite time series clustering is the choice of an appropriate distance measure, which can be used to analyze the similarities/dissimilarities of finite time series. Such a distance measure should be able to address questions such as: \textit{(i)} how to describe the underlying shape or trend in the time series? and \textit{(ii)} how to handle distortion in the amplitude and phase of the time series, such as noise or jitter?  
\\\indent Among different distance measurement methods that have been used for time series clustering, \textit{Euclidean distance} (ED) and \textit{Dynamic Time Warping} (DTW) are the most popular methods. In most of clustering techniques for time series data, Euclidean distance is used for measuring the similarities/dissimilarities of time series. However, this distance is most useful when we have time series in the form of equal length records. Euclidean distance in time series clustering only compares the points of the time series in a fixed order, so it cannot be used for clustering time series with time shifts. The DTW distance can be used for evaluating similarities/dissimilarities of time series data based on their shape information. However, this method is computationally expensive, since the cost of computing the similarity/dissimilarity between each pair of time series is quadratic in the length of the records.
\\\indent Most state-of-the-art approaches for shape-based clustering suffer from two main drawbacks: \textit{(i)} they are computationally expensive and thus, not suitable for large volumes of data \cite{kshape53,kshape59, kshape64, kshape87}; and \textit{(ii)} these approaches are limited to specific domains \cite{kshape87}, or their effectiveness has only been evaluated over a small number of datasets \cite{kshape53, kshape59}.
\\\indent k-Shape \cite{bibi14} is a novel method that has been proposed to address the problem of finding a suitable distance measure and clustering method for finite time series data, where time series sequences belong to the same cluster, exhibiting similar patterns, regardless of differences in amplitude and phase. Paparizos and Gravano \cite{bibi14} select a statistical measure, cross-correlation, as a shape similarity distance measure for comparing time series sequences. For clustering time series data, they modify the classical crisp k-means clustering algorithm (also called \textit{hard c-means} (HCM)) with a new centroid computation technique based on a shape extraction method while computing the clusters. These two novel modifications seem highly effective in terms of clustering accuracy and efficiency. To demonstrate the robustness of k-Shape clustering \cite{bibi14}, the authors perform an extensive evaluation over 48 different time series datasets and demonstrate the superiority of their technique against fifteen existing schemes.
\\\indent In this paper, we introduce two novel clustering algorithms. The first substitutes the \textit{Shape Based Distance} (SBD) measure introduced in \cite{bibi14} for the model norm in the standard fuzzy c-means \cite{bibi15} algorithm, and the second uses the SBD measure and shape extraction method in \cite{bibi14} to update prototypes in the fuzzy c-means clustering algorithm \cite{bibi15}. These two modifications result in $4.5\%$ and $58.5\%$ improvement respectively in accuracy and efficiency in comparison to k-Shape clustering. We make a direct comparison with the results reported in the k-Shape paper, by evaluating our methods on the same UCR time series classification archive \cite{bibi32}, which consists of 48 labeled time series datasets from various real-world domains.
\\\indent The rest of the paper is organized as follows: Section~\ref{sec:related} reviews related work on time series clustering, Section~\ref{sec:preli} presents the relevant theoretical background on fuzzy c-means and k-Shape clustering that are based of our two new algorithms. Section~\ref{sec:FCS++} introduces our two novel clustering methods FCS++ and FCS+. Section~\ref{sec:validity} discusses the four external cluster validity indices used for measuring the accuracy of clustering methods. Section~\ref{sec:experiment} evaluates our proposed methods with extensive numerical experiments, and shows that our algorithms outperform the crisp k-Shape clustering algorithm in terms of accuracy and efficiency. Sections~\ref{sec:discuss} and Section~\ref{sec:conclude} contain a short discussion of our conclusions and some future research directions.
%---------------------------------------------------------------------------------------------------------------
%                           Related work
%----------------------------------------------------------------------------------------------------------------\section{Rationale and Related Work}
\section{Related Work}\label{sec:related}
Among different techniques for analyzing time series data, clustering is a popular data mining method due to its usefulness as a preprocessing  step for other data mining techniques. There are comprehensive literature surveys on time series clustering such as\cite{kshape19,khodam1, jadid12}. Generally time series clustering algorithms fall in to two categories: Statistical-based and shape-based methods. Statistical-based methods, use characteristic measures \cite{kshape82}, varying coefficient models \cite{kshape38}, and subsequences of time series as approaches for extracting features to cluster time series data. However, there are two general drawbacks of using these methods: first, applying these methods usually leads to trial and error problems because of their requirement for adjusting multiple parameters; second, they are not domain-independent \cite{bibi14}. On the other hand, shape-based methods do not have these limitations and they have been used in a variety of studies. Therefore, in this paper we focus on shape-based clustering. In the following we briefly review shape-based distance functions for time series data as well as relevant clustering techniques that have been reported in the literature.
\\\indent An efficient shape-based distance function that measures the similarities/dissimilarities between time series data, should not be sensitive to the time of appearance of similar sequences \cite{jadid12}. Indeed, a distance function that measures the shape similarity should be invariant to phase and amplitude. Therefore, elastic methods \cite{jadid2,jadid3} like DTW \cite{jadid4} are more suitable for shape-based time series clustering.
\\\indent Next, we review relevant shape-based distance functions that have been introduced in previous studies. In \cite{jadid5} the authors introduced an indexing method using Euclidean distance in a lock-step measurement (one-to-one) manner. Sensitivity to scaling is the drawback of this method. The DTW distance has better accuracy but lower efficiency than this method. In \cite{kshape28} the authors proposed a cross-correlation based distance function that is very efficient for noise reduction. In addition, this distance function can summarize the temporal structure of the time series data. In \cite{jadid6,jadid7} a non-metric similarity measure is proposed based on the \textit{Longest Common Sub-Sequence} (LCSS), which has very good noise robustness. In this method, the similarity between trajectories is considered by giving more weight to similar segments of the sequences.
\\\indent In \cite{jadid8} the \textit{Edit Distance with Real Penalty} (ERP) is introduced as a metric that can handle local shifting in time series data. The ERP approach is a robust method against noise and is shift- and scale-invariant. In \cite{jadid9} the \textit{Minimal Variance Matching} (MVM) method is proposed which is able to skip outliers automatically. MVM uses the DTW method for calculating the similarity between two time series, but MVM, unlike DTW, can skip some target series. In addition, MVM, like the LCSS method, can allow the query sequences to match to only sub-sequence of the target sequence.
\\\indent In \cite{jadid10} the authors introduced the \textit{Edit Distance on Real sequence} (EDR) method, which is robust against noise, shifts, and scaling of the time series data. This method is more robust than the ED, DTW, and ERP methods, and it is outperforms the LCSS method. EDR measures the similarity between two trajectories based on the edit distance on strings. By assigning penalties to the unmatched segments, EDR improves its accuracy, and by quantizing the distance between 0 and 1 effectively removes the noise. In \cite{jadid11} the \textit{Sequence Weighted Alignment} (Swale) model is proposed to score similarities between time series data based on rewards for matching portions, and penalties for mismatching portion of sequences. \\\indent In \cite{kshape19,kshape81} the authors conducted extensive experiments on 38 data sets from a wide variety of application domains to compare distance functions. They concluded that the \textit{constrained Dynamic Time Warping} (cDTW) method \cite{bibi21} outperforms all other distance measurement methods in term of accuracy.
\\\indent The \textit{Shape Based Distance} (SBD) \cite{bibi14} is a novel measurement approach that handles distortions in amplitude and phase accurately, by efficiently using cross-correlation through normalization of a shape-based distance measure. According to the authors of \cite{bibi14}, SBD achieves similar accuracy to cDTW but it does not require any tuning, and it is an order of magnitude faster. In this paper we use SBD as the distance function in our algorithms.
\\\indent Many studies have focused on partitional models of clustering using a shape-based distance function that is shift- and scale-invariant. Among partitional clustering methods, k-medoids \cite{kshape40} clustering has been used widely because of its ease of incorporating a shape-based distance measure \cite{kshape19,kshape81}. However, k-medoids is a computationally expensive and non-scalable method. Alternatively, other studies \cite{kshape53,kshape59,kshape64,kshape87} have focused on k-means (aka hard c-means (HCM)) \cite{kshape50} with various choices of centroid computation and/or distance measure. The best performing k-means approaches have used DTW while proposing a new centroid extraction method\cite{kshape53,kshape59,kshape64,kshape87,bibi14}. 
\\\indent Recently, a novel partitional clustering algorithm, k-Shape, that uses the SBD distance, was proposed in \cite{bibi14}. The k-Shape algorithm preserves the shapes of time series when comparing them, and computes centroids in a scale- and phase- invariant way. The authors of \cite{bibi14} showed that k-Shape partitioning of 48 finite, labeled time series data sets was superior to 15 other methods based on spectral, partitional, and hierarchical clustering approaches. k-Shape appears to be a scalable and accurate approach for time series clustering that can be applied to a wide variety of domains.
\\\indent There are also fuzzy approaches to clustering in time series based on algorithms such as FCM. Studies based on fuzzy clustering of time series data include \cite{survey28, kshape28, khodam2}. In \cite{survey28} a \textit{short time series} (STS) distance is proposed to measure similarities between time series that are short in length,  based on their shape. By changing the Euclidean distance to STS distance in the standard fuzzy c-means algorithm, they modified the membership matrix and centroid computation algorithm, thus producing a fuzzy time series clustering algorithm for short time series data. In \cite{kshape28} the cross-correlation method is considered as the distance measure in the fuzzy c-means clustering algorithm and arithmetic means are used for prototype computation to cluster functional MRI data. The authors in \cite{jadid1} used the DTW method as distance measure and applied k-means and c-medoids algorithms as clustering methods. Comparing the results of these two methods, led them to conclude that k-means cannot produce satisfactory results whereas c-medoids obtained good accuracy. In this paper, we propose two novel shape-based clustering methods of time series data based on heuristic derivatives of FCM.
%------------------------------------------------------------------------------------------------
%             Preliminaries
%------------------------------------------------------------------------------------------------
\section{Preliminaries}\label{sec:preli}
Many clustering algorithms exist for numerical feature vector data such as $X=\{x_{1},x_{2},\dots,x_{n}\} \subset \Re^{p}$. There are three basic problems associated with clustering in $X :$ $(i)$ pre-clustering assessment of tendency (what value of c, the number of clusters, to use?); $(ii)$ partitioning the data (finding the c clusters); and $(iii)$ post-clustering validation (are the clusters useful, realistic, etc.?). Good presentations on each of the three basic problems include the general texts \cite{bibi1,bibi2,bibi3,bibi4}. This study concerns itself with three algorithms
for clustering in data where each $x_{k}\in X$ is a vector of length $p$ that represents a finite portion of a waveform or time series. Data of this type are prevalent in many fields: for example, stock market analysis \cite{bibi5}, satellite imagery \cite{bibi6}, computational finance \cite{bibi7} and neuroscience, \cite{bibi8,bibi9,bibi10}.
\\\indent In this section we review the relevant theoretical background for our new clustering algorithms. In Section ~\ref{sec:FCM}, we review fuzzy c-means clustering, which is the basis of our work. In Section ~\ref{sec:SBD}, we introduce the shape-based distance (SBD) \cite{bibi14}. Section ~\ref{sec:SE} introduces a new method for assigning a center to clusters based on the shape of the time series and briefly explains the k-Shape clustering method\cite{bibi14}.
%-----------------------------------------------------------------------------------------------
 %                  The Generalized Fuzzy C-Means Model
%-----------------------------------------------------------------------------------------------
\subsection{ The Generalized Fuzzy c-Means Model}\label{sec:FCM}
Crisp and fuzzy/probabilistic c-partitions of $n$ objects are, respectively, the sets of matrices defined as
%----------------------------------------------------------------------------------
%             equation 1
%----------------------------------------------------------------------------------
\begin{subequations} \label{eq:fcm1}
\begin{align} \label{eq:fcm1a}
&M_{fcn}=\{U \in \Re^{cn} :for\, 1 \leq i \leq c, 1 \leq k \leq n; 0 \leq u_{ik} \leq 1 ;\nonumber\\
&\qquad \displaystyle\sum_{i=1}^{c} u_{ik} = 1 \,\, \forall k; \, \displaystyle\sum_{k=1}^{n} u_{ik} > 0 \,\,\forall i\} \\
&\qquad M_{hcn}=\{U \in M_{fcn} : u_{ik} \in \{0,1\} \forall i,k\}.\label{eq:fcm1b}
\end{align}
\end{subequations}
%-----------------------------------------------------------------------------------------
The \textit{Generalized Least Squares Error} (GLSE) clustering model is the constrained optimization problem
%-------------------------------------------------------------------------------------
%                   equation2
%----------------------------------------------------------------------------------------
\begin{align} \label{eq:fcm2}
min_{(U,P)} \{J_{m}(U,P;X) = \displaystyle\sum_{k=1}^{n}\displaystyle\sum_{i=1}^{c}(u_{ik})^{m}[\Delta(x_{k},p_{i})]^{2} \}.
\end{align}
%-----------------------------------------------------------------------------------------
In equation \eqref{eq:fcm2}, $m$ is a weighting parameter, $m\geq1$, $U \in M_{fcn}$, $P=\{p_{1},\dots,p_{c}\}$ is a set of cluster prototypes, and $[\Delta(x_{k},p_{i})]^{2}$ is a measure of the squared error incurred by representing input $x_{k}$ by prototype $p_{i}$. Optimal partitions $U^{*}$ of $X$ are taken from pairs $(U^{*},V^{*})$ that are local minimizers of $J_{m}$. 
\\\indent There are dozens of instances of equation \eqref{eq:fcm2} for specific choices of $P$ and $\Delta$. For example, the prototypes $P$ may be points, lines, planes, linear varieties \cite{bibi3}, hyperquadrics \cite{bibi11}, or even regression functions \cite{bibi12}. The dissimilarity measure $\Delta$ may be an inner product norm \cite{bibi3}, a Minkowski norm \cite{bibi13}, or a shape-based distance \cite{bibi14} such as the one used in this paper.
\\\indent For many choices of $(P,\Delta)$, there is a theory to guide approximate solutions of the model in equation \eqref{eq:fcm2} by an \textit{alternating optimization} $(AO)$ algorithm based on Picard iteration through necessary conditions for its local extrema. The most familiar case is the original fuzzy c-means (FCM) model with accompanying FCM/AO algorithm, first reported in \cite{bibi15} and described at length in \cite{bibi3}. This case is covered by:\\
\\\indent \textbf{Theorem (1) (FCM)}\cite{bibi15}: Let $X=\{x_{1},\dots,x_{n}\} \subset \Re ^{p}$ contain at least $c$ distinct points. Let $P = V =\{v_{1},\dots,v_{c}\} \subset \Re ^{p}$, and $A$ be a positive-definite norm inducing $p \times p$ weight matrix, $\Delta(x_{k},p_{i})^{2} =\|x_{k} -v_{i}\|_{A}^{2} =(x_{k} -v_{i})^{T} A(x_{k} -v_{i}) >0 \quad\forall i,k$. If $m >1$, then $(U,V) \in M_{fcn} \times \Re ^{p}$ may minimize $J_{m}(U,V;X)$ only if\\
%-----------------------------------------------------------------------------------------------
%          equation 3
%------------------------------------------------------------------------------------------------
\begin{subequations} \label{eq:fcm3}
	\begin{align} \label{eq:fcm3a}
& u_{ik} =\left(\displaystyle\sum_{j=1}^{c} \left(\frac{\|x_{k} -v_{i}\|_{A}}{\|x_{k} -v_{j}\|_{A}}\right)^{\frac{2}{m-1}}\right)^{-1}; 1\leq i \leq c;1\leq k \leq n,  \\
& v_{i} =\displaystyle\sum_{k=1}^{n}u_{ik}^{m} x_{k} / \displaystyle\sum_{k=1}^{n}u_{ik}^{m}; \, 1\leq i \leq c.\label{eq:fcm3b}
	\end{align}
\end{subequations}
%-----------------------------------------------------------------------------------------
It is also well known that when $m = 1$ the partition matrix $U$ is necessarily crisp, $U\in M_{hcn}$, and under the remaining hypotheses in Theorem (1), that equation \eqref{eq:fcm2} reduces to the classical hard c-means (HCM, or classical k-means) model which was first discussed by Lloyd in \cite{bibi16}. In this case, there are several ways to show that conditions \eqref{eq:fcm3}  reduce to the following well known necessary conditions for minimizing $J_{1}$.\\
\\\indent\textbf{Theorem (2) (HCM)} \cite{bibi15} : Let $ X=\{x_{1},\dots,x_{n}\} \subset \Re^{p}$ contain at least $c$ distinct points, $P =V =\{v_{1},\dots,v_{c}\} \subset \Re ^{cp}$ , and $m=1$. If $\Delta (x_{k},p_{i})^{2} =\|x_{k} -v_{i}\|_{A}^{2} >0 \quad \forall i,k$, then $(U,V)\in M_{fcn} \times \Re^{cp}$ may minimize $J_{1}(U,V;X)$ only if $U\in M_{hcn}$ is a \textit{hard c-partition} of $X$, and
%--------------------------------------------------------------------------------------------
%                        equation4
%---------------------------------------------------------------------------------------------
\begin{subequations} \label{eq:fcm4}
  \begin{align} \label{eq:fcm4a}
& u_{ik} =\begin{Bmatrix}
     1 & \quad \|x_{k} -v_{i}\|_{A} < \|x_{k} -v_{j}\|_{A} \forall j\neq i\\
      0 & \quad \text{otherwise }\\
     \end{Bmatrix}\\
 & v_{i} = \displaystyle \sum_{k=1}^{n}u_{ik} x_{k} / \displaystyle \sum_{k=1}^{n}u_{ik};\, 1\leq i \leq c \,\, \label{eq:fcm4b}
 \end{align}
 \end{subequations}
%-----------------------------------------------------------------------------------------------
The content of Theorem (2) is often given by denoting the partition $U \in M_{hcn}$ by its equivalent set-theoretic form,\\ $X = \bigcup_{i=1}^{c}X_{i}$; $\oslash = X_{i}\bigcap_{i\neq j} X_{j}$; $\displaystyle\sum_{i=1}^{c} \mid X_{i}\mid= \displaystyle\sum_{i=1}^{c}n_{i} =n$. Using this representation, equations \eqref{eq:fcm4} take the alternate form
%------------------------------------------------------------------------------------------
%              equation 5
%-------------------------------------------------------------------------------------------
\begin{subequations} \label{eq:fcm5}
	\begin{align} \label{eq:fcm5a}
	& u_{ik} =\begin{Bmatrix}
	1 & \quad x_{k} \in X_{i}\\
	0 & \quad \text{otherwise }\\
	\end{Bmatrix} ; 1 \leq i \leq c , \, 1 \leq k \leq n\\
	& v_{i} =\bar{v}_{i} =\displaystyle\sum_{x_{k}\in X_{i}} \left(\frac{x_{k}}{n_{i}}\right) ; 1 \leq i \leq c.\label{eq:fcm5b}
	\end{align}
\end{subequations}
%-------------------------------------------------------------------------
The advantage of using equations \eqref{eq:fcm4} instead of \eqref{eq:fcm5} is that the role of the partition matrix $U$ is clearly seen in the more general fuzzy case. Note that \eqref{eq:fcm5a} labels each point in $X$ with the nearest prototype rule; and \eqref{eq:fcm5b} shows that the point prototypes are none other than the geometric centroids (sample means) of the $c$ clusters in $X$.
\\\indent There are various ways to estimate solutions for the GLSE problem \eqref{eq:fcm2} for the choice $[\Delta(x_{k},p_{i})]^{2} = \|x_{k}-v_{i} \|_{A}^{2}$. The most popular method is Picard iteration through necessary conditions \eqref{eq:fcm3} or \eqref{eq:fcm4}. For the fuzzy case, these are \textit{first order necessary conditions} (FONCs, the gradient must vanish at extreme points). For the crisp case, looping through conditions \eqref{eq:fcm4} or \eqref{eq:fcm5} is sometimes called Lloyd iteration: \eqref{eq:fcm4b}, \eqref{eq:fcm5b} is a FONC, and \eqref{eq:fcm4a}, \eqref{eq:fcm5a} is necessary, but not first order. This method is summarized as Algorithm~\ref{alg:A1}.
%------------------------------------------------------------------------------------------
%                         Algorithm1: FCM
%------------------------------------------------------------------------------------------
\renewcommand{\algorithmicrequire}{\textbf{Input:}}
\renewcommand{\algorithmicensure}{\textbf{Output:}}
\begin{algorithm}                      % enter the algorithm environment
\caption{AO c-means for HCM and FCM}          % give the algorithm a caption
\label{alg:A1} 
% and a label for \ref{} commands later in the document
\begin{algorithmic}[1]                   % enter the algorithmic environment
   
    \REQUIRE \quad $X=\{x_{1},\dots,x_{n}\} \subset \Re ^{p}$
    \ENSURE $(U,V)$
    \STATE Pick \quad$1<c<n : m\geq 1$ $T_{M}= iterationLimit$
    \STATE Model Norm:\quad $\parallel x-v\parallel_{A}= \sqrt{(x-v)^{T} A(x-v)}$
    \STATE $E_{t}=\mid J_{m}(U_{t},V_{t}) - J_{m}(U_{t-1},V_{t-1})\mid$
    \STATE $0<\varepsilon = terminationCriteria$
    \STATE Guess \quad $V_{0}=\{v_{10},\dots,v_{c0}\} \subset \Re ^{cp}$ 
    \STATE Calculate $U_{0}$ with $V_{0}$ and (3a) or (4a)
    \STATE $t=1$: $E_{1}= big Number$ 
    \WHILE{$t<T_{m} \quad and \quad E_{t}>\varepsilon$}
          \STATE Calculate $U_{t}$ with $V_{t-1}$ and (3a) or (4a)
          \STATE Calculate $V_{t}$ with (3b) or (4b)
          \STATE $t=t+1$
    \ENDWHILE
    \STATE $(U,V) \gets (U_{t},V_{t})$
\end{algorithmic}
\end{algorithm}
%--------------------------------------------------------------------------------------------
\\\indent Reference \cite{bibi3} contains details about most of the issues that arise concerning the use of these two clustering models and Algorithm~\ref{alg:A1}. Our interest is confined to three algorithms for clustering finite time series in the special case when each $x_{k} \in X$ is a vector of length $p$ that represents a finite portion of a waveform or time series. All three algorithms are ad hoc methods that depend on equations \eqref{eq:fcm4} or \eqref{eq:fcm5} and use Algorithm~\ref{alg:A1} with appropriate modifications to find approximate solutions.
%-------------------------------------------------------------------------------------------
%                    Shape based dictance
%-------------------------------------------------------------------------------------------
 \subsection{ Shape Based Distance}\label{sec:SBD}
The quality of clustering in feature vector data almost always depends on finding a good way to measure similarity or distance between the items represented by the data. Waveform data presents some special challenges, when represented by fixed length vectors of finite time series. Most of the extant literature on clustering in time series data modifies a classic clustering algorithm such as k-means by \textit{(i)} substituting a suitable distance measure on the raw data; or \textit{(ii)} by extracting features from the data that convert it into vectors that can be inserted directly into a classic algorithm. This paper follows the approach taken in \cite{bibi14}, which concentrates attention on SBD as the distance of choice for waveform clustering. 
\\\indent The underlying desire in waveform clustering is to use a distance measure that capture shapes differences that are invariant to phase and amplitude changes in the data. Constrained dynamic time warping is often recommended for this job \cite{bibi21}. Forty six (46) such measures for time series data are discussed in \cite{bibi25}! This gives some idea of how long and hard the path towards a good waveform distance measure has been. The shape-based distance (SBD) measure introduced in \cite{bibi14} for time series data forms the basis for our two new algorithms.
\\\indent The new algorithms begin by modifying the GLSE model at \eqref{eq:fcm2} by introducing the SBD, which is tailored to the time series case to define $\Delta(x_{k},p_{i})$. Specifically, this distance is introduced for a pair of z-normalized vectors $x$ and $y$ of length $p$ in Algorithm~\ref{alg:A2}:
\\
%--------------------------------------------------------------------------------
%                      Algorithm 2
%--------------------------------------------------------------------------------
\renewcommand{\algorithmicrequire}{\textbf{Input:}}
\renewcommand{\algorithmicensure}{\textbf{Output:}}
\begin{algorithm}                      % enter the algorithm environment
\caption{\cite{bibi14}: $[dist, y']= SBD(x,y)$}          % give the algorithm a caption
\label{alg:A2} 
% and a label for \ref{} commands later in the document
\begin{algorithmic}[1]                   % enter the algorithmic environment
   
    \REQUIRE Two z-normalized sequences $x,y \in \Re ^{p}$
    \ENSURE Dissimilarity dist = SBD(x,y) $\in \Re^{+}$;Aligned sequence y' of y towards x
    \STATE length = $2^{nextpower2(2*length (x)-1)}$
    \STATE $CC=IFFT\{FFT(x,length)^{*} FFT(y,length)\}$\%(A)
    \STATE $NCC_{c}=CC/\parallel x \parallel\bullet \parallel y \parallel$
    \STATE [value,index] = max($NCC_{c}$)
    \STATE dist= 1 - value \qquad \qquad \qquad \qquad \qquad \qquad \qquad  \% (B)
    \STATE shift = index - length(x)
     \IF{$shift \geq 0$}
        \STATE y'= [zeros(1,shift), y(1:end-shift)] \qquad \quad \qquad \%(C)
    \ELSE
        \STATE y'= [y(1-shift:end), zeros(1,-shift)] \qquad \qquad \quad  \%(C)
    \ENDIF
\end{algorithmic}
\end{algorithm}
%----------------------------------------------------------------------
%-----------------------------------------------------------------------------------
\\\indent \textbf{Notes about Algorithm~\ref{alg:A2}} The z-normalization mentioned in the input line is the standard $(0,1)$ statistical normalization, i.e., the $n$ input vectors are linearly transformed by subtracting their feature means and dividing by their standard deviations for each of the $p$ features in the input data. Lines $2$, $5$, and $8$ correspond to the following equations from \cite{bibi14}.
\\$(A)$ $CC(X,Y)= \mathcal{F} ^{-1} \{\mathcal{F}(x)^{*} \mathcal{F} (y) \}$, where $\mathcal{F}$ and $\mathcal{F}^{-1}$ are forward (inverse) discrete Fourier transforms, $*$ is convolution.
\\$(B)$ $dist= SBD (x,y)=1 -max_{w}\{CC_{w}(x,y)/ \|x\|\bullet \|y\|\}$, where $w \in \{1,2,\dots,2p-1\}$; $CC_{w}(x,y)=R_{w-p}(x,y)$; and \\ \\ $R_{k}(x,y)=\begin{Bmatrix}
\displaystyle\sum_{j=1}^{p-k} x_{j+k}y_{j}; & \quad k\geq 0\\
R_{-k}(x,y); & \quad k<0\\
\end{Bmatrix}$.
\\ $(C)$ $x_{(s)} =\begin{Bmatrix} ( \overbrace{0,\dots,0}^{\mid s \mid},x_{1},\dots,x_{p-s});
 & \quad s\geq 0\\
(x_{1-s},\dots,x_{p},\underbrace{0,\dots,0}_{\mid s \mid}); & \quad s<0\\
\end{Bmatrix}$.\\
\\\indent Equation $(C)$ shows how a sliding window passes across a vector, while looking for the optimal alignment between $x$ and $y$. The inputs to Algorithm~\ref{alg:A2} are vectors in $\Re^{p}$, so the distance $\|x_{k}-v_{i}\|_{A}$ in equations \eqref{eq:fcm4a} and \eqref{eq:fcm5a} can be directly replaced by $SBD(x,y)$, which is called a distance in \cite{bibi14}. Since $SBD(x,y)$ is produced by Algorithm~\ref{alg:A2}, it is not immediately obvious whether the function $\Delta = SBD$ is really a metric on $\Re^{p}$. Moreover, it is not clear whether the distance emerging from Algorithm~\ref{alg:A2} should be regarded as a squared distance or not, although the authors of \cite{bibi14} state that it replaces $[\Delta(x,v)]^{2}$ in the GLSE problem. The authors of \cite{bibi14} make this substitution in \eqref{eq:fcm4a} and in line 3 of Algorithm~\ref{alg:A1} with $m=1$, and call the resultant algorithm the $k-AVG+ED$ algorithm, where $AVG$ means \eqref{eq:fcm4b} or \eqref{eq:fcm5b} is used to compute the prototype updates, and $ED$ stands for Euclidean distance.
\\\indent Paparrizos and Gravano argue in \cite{bibi14} that most of the research on waveform clustering has used distances like DTW to the exclusion of \textit{cross correlation} $(CC)$ between time series because this time honored statistical method suffers from normalization and registration issues. They assert that to be useful, the data and the $CC$ applied to it must be appropriately normalized. The z-normalization of the inputs to Algorithm~\ref{alg:A2} gives it scale invariance, and takes care of inherent distortion in the data. They tackle the normalization of $CC$ with equations (A)-(C) that follow Algorithm~\ref{alg:A2}. Shift invariance is addressed by computing $CC_{w}(x,y)=R_{w-p}(x,y)$, which maximizes $CC$ at the best match between $x$ and $y$ by testing each position offered by the sliding window in equation (C). They discuss three ways to normalize $CC$, but only use $CC_{w}$, shown and used here, which produces values in the closed interval $[-1,1]$. Consequently, $SBD(x,y) \in [0,2]$, and it takes the value $0$ when $x$ and $y$ are perfectly similar (corresponding to zero distance, i.e., $x=y$). Efficiency of finding $CC_{w}$ is addressed by using the discrete forward and inverse Fourier transforms and padding the input data so its length is always a power of $2$. The overall complexity of Algorithm~\ref{alg:A2} is given as $O(plog(p))$.
%-----------------------------------------------------------------------------------------
%                       section shape based prototype
%-----------------------------------------------------------------------------------------
\subsection{ Shape Based Prototypes and k-Shape Clustering}\label{sec:SE}
The second major alteration of k-means introduced in \cite{bibi14} is to compute prototypes that attempt to capture shape information, based on the fact that input vectors represent finite time series waveforms. To begin, we revisit equation \eqref{eq:fcm3a} for $v_{i}$. This prototype arises for FCM by zeroing the gradient of the function $J_{m}(U^{*}, V; X)$ in the reduced optimization problem
 %-----------------------------------------------------------------------------------
 %                            equation 6
 %---------------------------------------------------------------------------------------
\begin{align} \label{eq:fcm6}
 min_{V \in \Re ^{cp}} \left\{J_{m}(U^{*}, V; X)= \displaystyle\sum_{k=1}^{n} \displaystyle\sum_{i=1}^{c}(u_{ik}^{*})^{m}\|x_{k}-v_{i}\|_{A}^{2}\right\}.
\end{align}
 %-------------------------------------------------------------------------------------------
$U^{*}$ is fixed in \eqref{eq:fcm6}, and minimization of $J_{m}(U^{*}, V; X)$ is unconstrained, so solving $\nabla _{v_{i}}J_{m}(U^{*}, V; X)=0$ for $v_{i}$ leads directly to \eqref{eq:fcm3a} in the fuzzy case, and \eqref{eq:fcm4a} in the crisp case. When the distance in equation \eqref{eq:fcm6} is not an inner product norm, $J_{m}(U^{*}, V; X)$ is generally not differentiable with respect to $v_{i}$. In this more general case, the reduced problem for the crisp case ($m=1$ in \eqref{eq:fcm6}) of the GLSE model becomes:
%------------------------------------------------------------------------------
%                       equation 7
%-----------------------------------------------------------------------------
\begin{align} \label{eq:fcm7}
min_{v \in \Re ^{p}} \left\{J_{1}(v; X)= \displaystyle\sum_{x_{k} \in X_{i}} \Delta(x_{k}-v)^{2}\right\}.
\end{align}
%--------------------------------------------------------------------------------------
The method of solving \eqref{eq:fcm7}, sometimes referred to as the Steiner sequence problem \cite{bibi18}, depends on the nature of the distance function $\Delta(x_{k}-v)$. When alignment of the observations $\{x_{k}\}$ with prototype $v$ is required, this becomes the multiple sequence alignment problem, known to be NP-complete \cite{bibi19}. Bypassing some intermediate steps given in \cite{bibi14}, \eqref{eq:fcm7} eventually becomes an instance of maximization of the famous Rayleigh Quotient \cite{bibi1}, viz,
%------------------------------------------------------------------------------------
%                     equation 8
%---------------------------------------------------------------------------------------
\begin{align} \label{eq:fcm8}
v_{i} = max_{v\in \Re^{p}}\left\{\frac{v^{T}Q^{T}SQV}{v^{T}v}\right\}= max_{v\in \Re^{p}}\left\{\frac{v^{T}Mv}{v^{T}v}\right\}
\end{align}
%-------------------------------------------------------------------------------------
where $M=Q^{T}SQ$, $Q=I-(1/p)[$\textbf{1}$]$, and $[$\textbf{1}$]$ is the $p\times p$ matrix of $1's$. The solution of \eqref{eq:fcm8} is the eigenvector  corresponding to the largest eigenvalue of the real symmetric matrix $M$, which is computed in the last line of Algorithm~\ref{alg:A3}, which records this procedure as shown in \cite{bibi14}.
%----------------------------------------------------------------------------------------
%                      Algorithm 3: shape extraction
%--------------------------------------------------------------------------------------
\renewcommand{\algorithmicrequire}{\textbf{Input:}}
\renewcommand{\algorithmicensure}{\textbf{Output:}}
\begin{algorithm}                      % enter the algorithm environment
\caption{\cite{bibi14}: $V_{i,t+1}$= \textbf{Shape Extraction}($X,v_{i,t}$)}          % give the algorithm a caption
\label{alg:A3} 
% and a label for \ref{} commands later in the document
\begin{algorithmic}[1]                   % enter the algorithmic environment
   
    \REQUIRE $X=\{x_{1},\dots,x_{n}\} \subset \Re ^{p}$ as a $n \times p$ matrix $X$ whose columns are z-normalized time series vectors. $v_{i,t}\in \Re^{p}$ is the reference sequence against which time series of $X$ are aligned 
    \ENSURE $v_{i,t}\in \Re^{p}$ \qquad \qquad \quad \% new $i^{th}$ shape prototype
    \STATE $X'\leftarrow$ $[\quad]$
    \STATE \textbf{for} $i=1$ \textbf{to} $n$ \textbf{do}
    \STATE \qquad [dist,$x'$] $\leftarrow$ SBD($v_{i,t},X(i)$) \quad \%Algorithm~\ref{alg:A2}
    \STATE \qquad $X'$ $\leftarrow$ $[X';x']$
    \STATE \textbf{end for}
    \STATE \qquad $S$ $\leftarrow$ $X'^{T} \bullet X'$ \qquad \qquad \qquad \%$S$, Eq (8)
    \STATE \qquad $Q$ $\leftarrow$ $I - \frac{1}{p}$[\textbf{1}] \qquad \qquad\qquad \%$Q$, Eq (8)
    \STATE \qquad $M$ $\leftarrow$ $Q^{T}\bullet$ $S$ $ \bullet$ $Q$ \qquad \qquad \%$M$, Eq (8)
    \STATE \qquad $v_{i,t}$ $\leftarrow$ $Eig(M,1)$ \qquad \qquad \%Extract $1^{st}$ev 
\end{algorithmic}
\end{algorithm}
%------------------------------------------------------------------------------------------
\\\indent\textbf{Notes about Algorithm~\ref{alg:A3}.} Lines 6, 7 and 8 compute the quantities $S$, $Q$ and $M$ needed in equation \eqref{eq:fcm8} to realize line 9, which produces updated prototype  $v_{i,t+1}$ for cluster $i$ given the points $(X_{i})$ currently in it and the current prototype $v_{i,t}$. Since there are $c$ prototypes, the k-Shapes (of course $k=c$ here) algorithm in \cite{bibi14} returns to Algorithm~\ref{alg:A3}, $c$ times during the prototype refinement step of the k-Shape clustering iteration.
\\\indent Now everything is in place for Papparrizos and Gravano to define their k-Shape method, repeated here from \cite{bibi14}.
%---------------------------------------------------------------------------------------
%                        Algorithm 4 k-Shape
%--------------------------------------------------------------------------------------
\renewcommand{\algorithmicrequire}{\textbf{Input:}}
\renewcommand{\algorithmicensure}{\textbf{Output:}}
\begin{algorithm}                      % enter the algorithm environment
\caption{\cite{bibi14}: $[u,V]$= \textbf{k-Shape}($X,c$)}          % give the algorithm a caption
\label{alg:A4} 
% and a label for \ref{} commands later in the document
\begin{algorithmic}[1]                   % enter the algorithmic environment
   
    \REQUIRE $X=\{x_{1},\dots,x_{n}\} \subset \Re ^{p}$ as a $n \times p$ matrix $X$ whose columns are z-normalized time series vectors: $c$ is the number of clusters produced
    \ENSURE \textbf{u} is a $n \times 1$ label vector that partitions $X$ into $c$ crisp clusters. $V$ is a $c\times p$ matrix containing $c$ shape extracted centroids of length $p$.
    \STATE $iter \leftarrow 0$
    \STATE $u' \leftarrow [\quad]$
     \WHILE{$u!=u'$ and $iter < 100$}
          \STATE $u'\leftarrow u$
          \\\% \textit{Refinement step}
          \STATE \quad \textbf{for} $j \leftarrow 1$ to $c$ \textbf{do}
          \STATE  \quad \quad $X'\leftarrow [\quad]$
          \STATE \quad \quad \textbf{for} $i \leftarrow 1$ to $n$ \textbf{do} 
          \STATE \quad \quad \quad \textbf{if} u(i)= j \textbf{then}
          \STATE \quad \quad \quad \quad $X' \leftarrow [X', X(i)]$
          \STATE \quad \quad \quad \textbf{end if}
          \STATE \quad \quad \textbf{end for}
          \STATE \quad $v(j) \leftarrow$ Shape Extraction $(X'; v(j))$ \%Algorithm3
          \STATE \quad \textbf{end for}
          \\ \% \textit{Assignment step}
          \STATE \quad \textbf{for} $i \leftarrow 1$ to $n$ \textbf{do}
          \STATE \quad \quad $mindist \leftarrow \infty$\\
          \STATE \quad \quad \textbf{for} $j \leftarrow 1$ to $c$ \textbf{do}
          \STATE \quad \quad \quad $[dist,x'] \leftarrow$ SBD ($v(j),X(i)$) \%Algorithm2
          \STATE \quad \quad \quad \textbf{if} $dist < mindist$ \textbf{then}
          \STATE \quad \quad \quad \quad $mindist \leftarrow dist$
          \STATE \quad \quad \quad \quad $u(i) \leftarrow j$
          \STATE \quad \quad \textbf{end if}
          \STATE \quad \quad \textbf{end for}
          \STATE \quad \textbf{end for}
          \STATE $iter \leftarrow iter+1$   
    \ENDWHILE    
\end{algorithmic}
\end{algorithm}
%----------------------------------------------------------------------------------------
%----------------------------------------------------------------------------------------
%-------------------------------------------------------------------------------------------
%            FCS+ AND FCS++ clustering section
%-------------------------------------------------------------------------------------
\section{FCS+ and FCS++ Clustering}\label{sec:FCS++}
The SBD algorithm, in conjunction with Theorem (1), offers a way to immediately modify the basic FCM algorithm for waveform inputs. We can replace the inner product norm in Line 2 of Algorithm~\ref{alg:A1} with the SBD computed by Algorithm~\ref{alg:A2}. This results in our first heuristic generalization, which we will call \textit{fuzzy c-Shape plus} (FCS+). We have added the "+" to the acronym FCS so that you won't confuse this new shape-based distance algorithm with an older one bearing the acronym FCS (\textit{fuzzy c-shells}, \cite{bibi17}), in which the prototypes are shell boundaries and the model norm is an inner product A-norm. 
%-------------------------------------------------------------------------------------
%                                 Algorithm 5 FCS+
%------------------------------------------------------------------------------------
\renewcommand{\algorithmicrequire}{\textbf{Input:}}
\renewcommand{\algorithmicensure}{\textbf{Output:}}
\begin{algorithm}                      % enter the algorithm environment
\caption{\textbf{FCS+}}          % give the algorithm a caption
\label{alg:A5} 
% and a label for \ref{} commands later in the document
\begin{algorithmic}[1]                   % enter the algorithmic environment
   \STATE \textbf{Replace}: line 3 in Algorithm FCM with: 3 Model Norm: SBD($x,v$) $\leftarrow \parallel x - v\parallel_{A}^{2}$
 \STATE \textbf{Do}: Algorithm 1
 \STATE $[optional]$ Harden $U$, line $13$, Algorithm 1 with Eq$(11)$
 \STATE $[optional]$ Output: $H_{mm}(U)\in M_{hcn}$
\end{algorithmic}
\end{algorithm}
%-----------------------------------------------------------------------------
\\\indent\textbf{Note about Algorithm~\ref{alg:A5}}. The optional lines of Algorithm~\ref{alg:A5} harden the terminal fuzzy partition. This option is NOT necessary, since there are soft versions of all four CVIs based on the contingency matrix discussed in the next section \cite{bibi29}. However, we will use this option for FCS+ in our numerical experiments so that the comparison of FCS+ to k-Shape (which has only crisp partitioning) is equitable. 
\indent Algorithm~\ref{alg:A4} represents the crisp partition of $X$ that it produces as the $n \times 1$ label vector \textbf{u} (not bold in Algorithm~\ref{alg:A4}). This is an efficient way to carry the information in the crisp case. To see how a fuzzy generalization would make sense, let us represent the partition information possessed by the crisp vector \textbf{u} as a matrix $U \in M_{hcn}$. For example, suppose the output of Algorithm~\ref{alg:A4} is $ \textbf{u}=[12131]$, so there are $c = 3$ crisp labels for $n=5$ objects. The matrix representation for this \textbf{u} is\\
%------------------------------------------------------------------------------------
%                    equation 9
%---------------------------------------------------------------------------------------
\begin{align} \label{eq:fcm9}
U =
\begin{bmatrix}
1&0&1&0&1 \\
0&1&0&0&0 \\
0&0&0&1&0\\
\end{bmatrix}
\end{align}\\
%---------------------------------------------------------------------------------
During the refinement step, Algorithm~\ref{alg:A4} will update the three cluster centers by calling Algorithm~\ref{alg:A3} three times. Each prototype update uses only the data vectors in its cluster, illustrated graphically as follows:\\
%---------------------------------------------------------------------------------------
%                            equation 10
%--------------------------------------------------------------------------------------
\begin{align} \label{eq:fcm10}
U =
\begin{bmatrix}
1&0&1&0&1 \\
0&1&0&0&0 \\
0&0&0&1&0\\
\end{bmatrix}
\quad
\begin{matrix}
\implies&v_{1,t}&\to&v_{1,t+1} \\
\implies&v_{2,t}&\to&v_{2,t+1} \\
\implies&v_{3,t}&\to&v_{3,t+1} \\
\end{matrix}
\end{align}\\
%-----------------------------------------------------------------------------------------
Equation \eqref{eq:fcm10} shows how \textit{crisp} membership in each cluster is used to control which vectors in the data set are accessed during the update procedure. At each $j$, lines 6-9 of Algorithm~\ref{alg:A4} picks out only the vectors in current cluster $X_{j}$ (corresponding to the 1's in the $j^{th}$ row of $U$) and writes them into $X^{'}$, so when Algorithm~\ref{alg:A3} is called in line 12 of Algorithm~\ref{alg:A4}, only the points in $X_{j}$ are sent to it via array $X^{'}$ to update the shape based prototype for the $j^{th}$ cluster.
\\\indent The representation of the partition produced by Algorithm~\ref{alg:A4} at \eqref{eq:fcm10} shows how to generalize the k-Shape algorithm to the fuzzy case. Algorithm FCS+ produces a fuzzy $U \in M_{fcn}$, which we can harden using the maximum membership function $\textbf{H}_{mm}: M_{fcn}\to M_{hcn}$, which operates on the columns of $U.$ $\textbf{H}_{mm}(U) =[\textbf{h}(\textbf{U}^{(1)})\cdots \textbf{h}(\textbf{U}^{(n)})]$ is a hardening of $U$ defined on the $n$ columns $\{\textbf{U}^{(k)}\}$ of $U$ as follows:
%----------------------------------------------------------------------------
%                         equation 11
%----------------------------------------------------------------------------
\begin{align} \label{eq:fcm11}
\textbf{h}(\textbf{U}^{(k)})=(0,0,\dots,\underbrace{1}_{i^{th}},\dots)^{T} \nonumber\\
   \iff \begin{Bmatrix} 1 & \quad u_{ik} > u_{jk} ; j \neq i\\
  0 & \quad \text{otherwise}\\
  \end{Bmatrix}.
\end{align}\\
In words: $\textbf{h}$ replaces the largest value in each column of $U$ with a $1$, and place $0$'s in the other $c-1$ slots in each column of $U$. When ties occur, assign the membership $1$ arbitrarily to any winner, and treat the other maximums as non-winners. For example if we apply $H_{mm}$ to the matrix\\
%------------------------------------------------------------------------------------
%                   equation 12
%------------------------------------------------------------------------------------
\begin{align} \label{eq:fcm12}
U' =
\begin{bmatrix}
0.9&0.1&0.6&0.3&0.65 \\
0.1&0.8&0.3&0.22&0\\
0&0.1&0.1&0.75&0.35\\
\end{bmatrix}
\end{align}\\
%---------------------------------------------------------------------------------------
the result is that $\textbf{H}_{mm}(U^{'})=U$, the matrix at \eqref{eq:fcm10}.
\\\indent Once this conversion is made, it is a simple matter to convert $\textbf{H}_{mm}(U^{'})$ back to the list form row by row, \textbf{u}\textbf{$\leftarrow$}$ \textbf{H}_{mm}(U^{'})$, required as one of the inputs to Algorithm~\ref{alg:A4}. With this conversion in hand, we can define the \textit{fuzzy c-Shape} double plus (FCS++) algorithm, where the first $+$ stands for the use of the SBD distance function (Algorithm~\ref{alg:A2}), and the second $+$ represents the use of SE prototypes via Algorithms~\ref{alg:A3} and ~\ref{alg:A4}.
%--------------------------------------------------------------------------------
%                   algorithm 6 FCS++
%-------------------------------------------------------------------------------
\renewcommand{\algorithmicrequire}{\textbf{Input:}}
\renewcommand{\algorithmicensure}{\textbf{Output:}}
\begin{algorithm}                      % enter the algorithm environment
\caption{\textbf{FCS++}}          % give the algorithm a caption
\label{alg:A6} 
% and a label for \ref{} commands later in the document
\begin{algorithmic}[1]                   % enter the algorithmic environment
    \REQUIRE $X=\{x_{1},\dots,x_{n}\} \subset \Re ^{p}$
    \ENSURE $(U,V)$
   \STATE \textbf{Pick} $1<c<n : m\geq 1 : T_{M}$= iteration limit
 \STATE Model Norm: $\Delta (x,v) =$ SBD($x,v$) 
 \STATE Error Norm: $E_{t} =$ $\parallel V_{t} - V_{t-1} \parallel$
 \STATE $0<\varepsilon =$ termination criterion
 \STATE \textbf{Guess} \quad $V_{0}=\{v_{10},\dots,v_{c0}\} \subset \Re ^{cp}$
 \STATE $t=1:$ $E_{1}=$ big number
 \WHILE{$T<T_{M}$ and $E_{t}>\varepsilon$}
 \STATE Calculate $U_{t}$ with $V_{t-1}$ and (3a)
 \STATE \textbf{u}$\leftarrow$ \textbf{H}($u_{t}$) \qquad \qquad \qquad \%Harden $U_{t}$ with Eq(11) \\\%\textit{SE Refinement step}\\
 \STATE \textbf{for} $j\leftarrow 1$ to $c$ \textbf{do}
 \STATE \quad $X' \leftarrow [\quad]$
 \STATE \quad \textbf{for} $i\leftarrow 1$ to $n$ \textbf{do}
 \STATE \qquad \textbf{if} $u(i)=j$ \textbf{then}
 \STATE \qquad \quad $X'\leftarrow [X',X(i)]$
 \STATE \qquad \textbf{end if}
 \STATE \qquad $v(j)\leftarrow$ Shape Extraction ($X';v(j)$) \%Algorithm3
 \STATE \quad \textbf{end for}
 \STATE \textbf{end for}
 \STATE \quad $V_{t} = [V(1) V(2) \cdots v(c)]$
 \STATE \quad $t=t+1$
 \ENDWHILE
 \STATE $(U,V)\leftarrow (u_{t}, V_{t})$ \qquad \% $U\in M_{hcn}$ is crisp
\end{algorithmic}
\end{algorithm}
%--------------------------------------------------------------------------------------
%               Cluster validity indices
%-------------------------------------------------------------------------------------------
\section{Cluster Validity Indices}\label{sec:validity}
The quality of our experimental outputs can be judged in a number of ways. To make direct comparisons between the k-Shape outputs in \cite{bibi14} to those found by FCS+ and FCS++, we will follow \cite{bibi14} by using external \textit{cluster validity indices} (CVIs), which are functions that identify a "best" member amongst a set of \textit{candidate partitions} $CP=\{U\in M_{fcn} : c_{m} \leq c \leq c_{M}\}$ of any set of $O=\{o_{1},\dotsc,o_{n}\}$ of $n$ objects.
\\\indent There are two basic types of CVIs. \textit{Internal} CVIs use only the information available from the algorithmic outputs to assess the quality of each $U\in CP$. \textit{External} CVIs use the information available to internal indices, but also use "outside" information about the data, which almost always means that the data are labeled by a crisp ground truth partition. So, external CVIs basically compare partitions in CP obtained by a clustering algorithm to the crisp partition of ground truth labels. This is rightly regarded as "fake clustering," since the subsets in the labeled data are, presumably, already clustered (but please note that these are labeled subsets, which may or may not be regarded as clusters by a specific model and algorithm). But this is a good way to compare different clustering algorithms (which is our aim here), and so many studies of this kind exist in the literature. 
\\\indent Chapter $16$ in \cite{bibi1} is an excellent source of general information about CVIs. The seminal paper by Milligan and Cooper \cite{bibi28} was the first comprehensive study of the selection of an internal CVI. The use of external CVIs to choose a "good" internal CVI as discussed in \cite{bibi29, bibi30}. Many other internal CVIs that are popular for validation of crisp partitions are compared in \cite{bibi31}, which offers a wide choice of potential CVIs that might be adapted to the present application. Here we briefly summarize the four external indices we use to evaluate our algorithms.
\\\indent Let $U\in M_{hrn}$, $Q\in M_{hcn}$ and $N=UQ^{T}$ (in general, $r \neq c$). When a reference (ground truth) partition is available, it will be $Q$ in the transformation $N=UQ^{T}$. The matrix $N$ forms a contingency table between the two partitions, so is sometimes called a \textit{contingency matrix}. Given $Q$, we match $Q$ to candidate $U\in CP$ using $CVI(U\mid Q)$. The Rand index \cite{bibi34}, one of the first (and still most popular) crisp external CVIs, is based on \eqref{eq:fcm4} paired comparison values derived from the elements of $N$ :\\
%----------------------------------------------------------------------
%                     Equation 13
%---------------------------------------------------------------------
  \begin{subequations} \label{eq:fcm13}
  	\begin{align} \label{eq:fcm13a}
  	& a = \frac{1}{2}\displaystyle\sum_{i=1}^{r} \displaystyle\sum_{j=1}^{c} n_{ij}(n_{ij}-1);\\
  	& b =  \frac{1}{2}\left(\displaystyle\sum_{j=1}^{c} n_{\bullet j}^{2} - \displaystyle\sum_{i=1}^{r} \displaystyle\sum_{j=1}^{c} n_{ij}^{2}\right);\label{eq:fcm13b}\\
  	& c=  \frac{1}{2}\left(\displaystyle\sum_{i=1}^{r} n_{i\bullet}^{2} - \displaystyle\sum_{i=1}^{r} \displaystyle\sum_{j=1}^{c} n_{ij}^{2}\right);\label{eq:fcm13c}\\
  	& d=  \frac{1}{2}\left(n^{2} + \displaystyle\sum_{i=1}^{r} \displaystyle\sum_{j=1}^{c} n_{ij}^{2} - (\displaystyle\sum_{i=1}^{r} n_{i\bullet}^{2} + \displaystyle\sum_{j=1}^{c} n_{\bullet j}^{2}) \right).\label{eq:fcm13d}
  	\end{align}
  \end{subequations}
%---------------------------------------------------------------------------------------------
CVIs are notoriously fickle, so we use four CVIs here based on different combinations of the values in the contingency matrices to see if they will all rank the three algorithms the same way. The \textit{Rand Index} (RI) is computed with these four values as
%------------------------------------------------------------------------------------------
%          equation14
%----------------------------------------------------------------------------------------------
\begin{align} \label{eq:fcm14}
RI(U \mid Q)= \frac{(a+d)}{(a+d)+(b+c)}
\end{align}
%---------------------------------------------------------------------------------------------
The numerator $(a+d)$ is the number of $agreements$ between pairs in $U$ and $Q$; $(b+c)$ is the number of $disagreements$. The $RI$ is valued in $[0, 1]$, taking its maximum if and only if $U =Q$. So, the heuristic for this index is that the maximum value of the RI over the partitions in $CP$ points to the best match among the candidates. We call such an index a $max-optimal\quad CVI$, indicated as $(\uparrow)$. Several adjustments of \eqref{eq:fcm14} have been proposed in the literature that attempt to rectify the tendency of the RI to increase monotonically with $c$. Of these, the \textit{Adjusted Rand Index} (ARI) of Hubert and Arabie \cite{bibi35} is the most popular:
%---------------------------------------------------------------------------------------------
%                         equation 15
%---------------------------------------------------------------------------------------------
\begin{align} \label{eq:fcm15}
ARI(U \mid Q)= \frac{\left(a-\frac{(a+c)(a+b)}{a+b+c+d}\right)}{\left(\frac{(a+c)+(a+b)}{2}-\frac{(a+c)(a+b)}{a+b+c+d}\right)}.
\end{align}
%------------------------------------------------------------------------------- -----------------
The $ARI$ is also max-optimal $(\uparrow)$: it maximizes at $1$, but its minimum may be negative if the index is less than its expected value of zero. The third external CVI we will use is the \textit{variation of information} (VI) introduced in \cite{bibi36}:
%---------------------------------------------------------------------------------------------
%                                equation16
%---------------------------------------------------------------------------------------------
\begin{align} \label{eq:fcm16}
VI(U \mid Q)= &-\displaystyle\sum_{i=1}^{r}\displaystyle\sum_{j=1}^{c}\frac{n_{ij}}{n}log\frac{n_{ij}}{n}\nonumber\\
&-\left[\displaystyle\sum_{i=1}^{r}\displaystyle\sum_{j=1}^{c}\frac{n_{ij}}{n}log\frac{n_{ij}/n}{(n_{i\bullet}/n)(n_{\bullet j}/n)}\right].
\end{align}
%---------------------------------------------------------------------------------------------
In equation \eqref{eq:fcm16} $n_{i\bullet}$, $n_{\bullet j}$ are, respectively, the $i^{th}$ row and $j^{th}$ column sums of the contingency matrix $N$.
\\\indent The $VI$ index is a metric valued in $[0,log(n)]$ which takes the value $0$ when $U = Q$, so the heuristic for $VI$ is to accept the partition achieving the minimum value over $CP$: the $VI$ is min-optimal $(\downarrow)$. 
\\The last external $CVI$ we use is a form of mutual information, which is normalized by a maximum calculation, so this index bears the notation $NMI_{M}(U\mid Q)$:
%--------------------------------------------------------------------------------------------
%                  EQUATION17
%--------------------------------------------------------------------------------------------
\begin{align} \label{eq:fcm17}
NMI_{M}(U \mid Q)= &\frac{\displaystyle\sum_{i=1}^{r}\displaystyle\sum_{j=1}^{c}\frac{n_{ij}}{n}log\frac{n_{ij}/n}{(n_{i\bullet}/n)(n_{\bullet j}/n)}}{max\{H_{S}(U),H_{S}(Q)\}},
\end{align}
%------------------------------------------------------------------------------------------
where, for example, $n_{i}=\displaystyle\sum_{k=1}^{n}u_{ik}$ is the number of points in the $i^{th}$ cluster in $U$, and $H_{S}(U)= -\displaystyle\sum_{k=1}^{r}\displaystyle\sum_{i=1}^{c}\frac{n_{i}}{n}log\frac{n_{i}}{n}$ is Shannon's entropy of $U$ \cite{bibi1}, and likewise for $H_{S}(Q)$. This index has been a good performer in several cluster validity studies \cite{bibi37}, and provides a nice contrast to the other three indices. The range of $NMI_{M}(U\mid Q)$ is $[0,1]$, and it is a max-optimal $(\uparrow)$ $CVI$, so larger values point to better matches between $U$ and $Q$.
\quad\\\indent The four CVIs we use can all be generalized to compare soft $U$'s to crisp $Q$ by the method discussed in \cite{bibi29}. However, we will convert FCS+ outputs to the crisp partition $H_{mm}(U)$ per Equation ~(\ref{eq:fcm11}) before computing these indices. And we MUST convert $U$ to $H_{mm}(U)$ to implement FCS++. Consequently, all three clustering algorithms will be evaluated using the same type of information.
%----------------------------------------------------------------------------------------
%                       figure1   FCS++, FCS+, c-shapes plots
%---------------------------------------------------------------------------------------
\begin{figure*}[t!]%
	\centering
	\begin{subfigure}[t]{0.32\textwidth}
		\includegraphics[width=\textwidth]{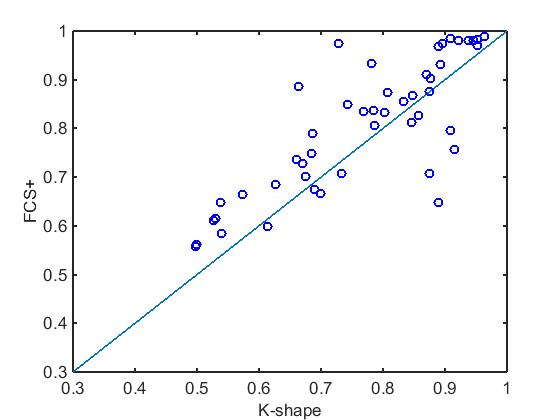}%
		\caption{FCS+ vs. k-Shape. Circles above the diagonal indicate datasets for which FCS+ has a better average Rand Index than k-Shape.}
		\label{fig:FCS+}
	\end{subfigure}\hfill%
	\begin{subfigure}[t]{0.32\textwidth}%\hspace{13mm}%
		\includegraphics[width=\textwidth]{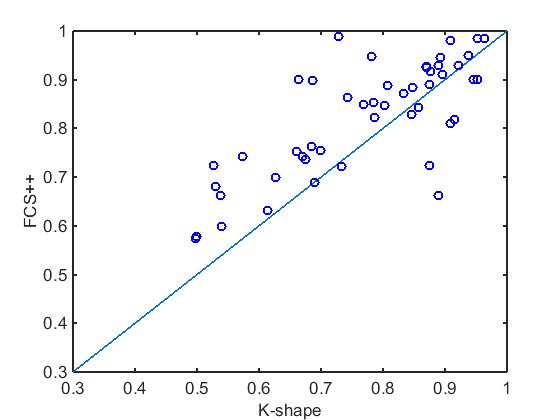}%
		\caption{FCS++ vs. k-Shape. Circles above the diagonal indicate datasets for which FCS++ has a better average Rand Index than k-Shape.}
		\label{fig:FCS++}
	\end{subfigure}\hfill%
	\begin{subfigure}[t]{0.32\textwidth}
		\includegraphics[width=\textwidth]{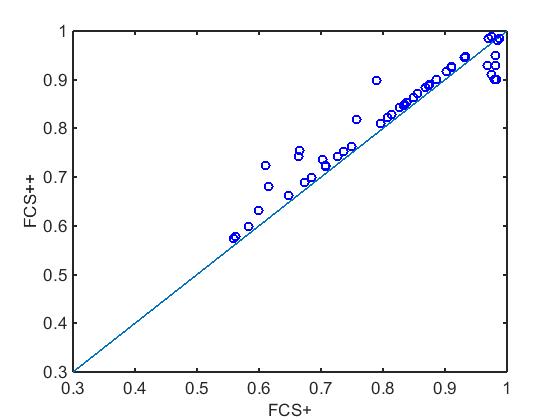}%
		\caption{FCS++ vs. FCS+. Circles above the diagonal indicate datasets for which FCS++ has better average Rand Index than FCS+.}
		\label{fig:TWOFCS}
	\end{subfigure}
	\caption{Comparisons of k-Shape, FCS+ and FCS++ :
		average RI values of ten runs over 48 data sets.
	}\label{fig:FCS}
\end{figure*}
%-------------------------------------------------------------------------------------------
%------------------------------------------------------------------------------------------------
%                    Experimental results Section
%------------------------------------------------------------------------------------------------
\section{Numerical Experiments}\label{sec:experiment}
\begin{figure}[h]
	\centering
	\includegraphics[width=.8\linewidth]{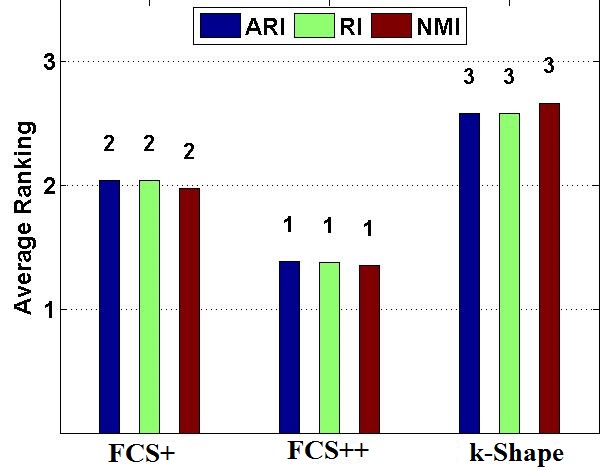}
    \caption{Comparison for rankings of clustering accuracy methods for 3 metrics. The bars represent average rankings based on the Friedman test, and the number on the top of the bars indicates the ranking of the algorithm, from the best (1) to the worst (3) for each given measure. The ranking is determined for all datasets and finally an average is calculated as the mean of all rankings. }%
		\label{fig:fridman}%
\end{figure}

We use the same 48 synthetic and real data sets that were used in \cite{bibi14}. The 48 sets are all z-normalized, crisply labeled, split into training and test sets, and are collected in the UCR time series database \cite{bibi32}. Our experiments are based on merged versions of the training and test sets. The length of the time series (number of points sampled from the waveforms) in each set is fixed and equal, ranging from $p = 24$ to $p = 1882$. The number of sample waveforms in the data sets ranges from $n = 56$ to $n = 9236$. Since all 48 of these data sets are class-labeled, we do not use the external CVIs to select a best candidate from a set of c-partitions at different values of $c$. Here we simply run FCS+ and FCS++ using the specified number of target labels for each of the 48 data sets, acknowledging that the number of labeled classes does not automatically identify the number of clusters that any model and algorithm might think are present in the data. We ran the experiments for each clustering method ten times for each data set. We implemented the clustering methods in MATLAB R2014b (64bits) using the platform: Intel(R) with core(TM)i7 processor and clock speed at 3.60 GHz and 16 GB RAM.
\\\indent Table~\ref{table:tabel1} shows the grand averages over $480$ trials for each of the four crisp external CVIs. Each index was evaluated 10 times for different runs of FCS+ and FCS++ on each of the $48$ labeled data sets. The ordering for the three algorithms is the same for all four indices. Table I shows that FCS++ is slightly better than FCS+, and in turn, FCS+ outperforms k-Shape by a small margin for all four indices. This demonstrates that the overall quality of both of the new fuzzy methods for clustering waveform data is superior to the k-Shape algorithm in \cite{bibi14}.
%-----------------------------------------------------------------------------------------
%                                    Table I
%-----------------------------------------------------------------------------------------
\begin{table}[ht]
\centering
\caption{Grand Average of (480) CVI values\\(10 runs per data set x 48 data sets)}% title of Tab
\begin{tabular*}{\linewidth}{@{\extracolsep{\fill}}c c c c c c}% centered columns (4 columns)
\toprule                       %inserts double horizontal lines
CVI & Type & Range & FCS++ & FCS+ & k-Shape \\[0.5ex]% inserts table
%heading
\midrule
$\overline{\overline{RI}}$& $(\uparrow)$& $[0,1]$ & \textbf{0.822}& 0.807& 0.772\\[1ex]% inserting body of the table
$\overline{\overline{ARI}}$ &$(\uparrow)$ & $[-a,1]$ & \textbf{0.461} &0.403 &0.321  \\[1ex]
$\overline{\overline{NMI_{M}}}$ & $(\uparrow)$& $[0,1]$ &\textbf{0.641} &0.534 & 0.413  \\[1ex]
$\overline{\overline{VI}}$ & $(\downarrow)$ & $[0,logn]$  &\textbf{1.010}&1.463 &2.455\\[1ex]% [1ex] adds vertical space
 \bottomrule %inserts single line
\end{tabular*}
\label{table:tabel1} % is usedto refer this table in the text
\end{table}
%----------------------------------------------------------------------------------------------
\\\indent Figure~\ref{fig:FCS} compares each pair of methods graphically. The line through the origin at $45$ degrees in each view separates $[0,1]\times [0,1]$ into two half-spaces. Each of the views ~\ref{fig:FCS+}, ~\ref{fig:FCS++} and ~\ref{fig:TWOFCS} plots the average values of the Rand Index over $10$ runs for each of the 48 data sets. The dots represent the $48$ data sets, and \textit{the coordinates of each dot are the average values of the Rand index for the algorithm pair in each view}.
\\For example, the coordinates of points in Figure~\ref{fig:FCS+} are: horizontal coordinate x = average RI value achieved by $10$ runs of k-Shape; vertical coordinate y = average RI value achieved by $10$ runs of FCS+. There are 10 points below the line y = x in Figure ~\ref{fig:FCS+}, 1 point on the line, and 37 points above the line. This means that FCS+ achieved a better average result with the RI than k-Shape on 37 of the 48 data sets, they were tied on one data set, and k-Shape had a better average RI than FCS+ on 10 data sets. Figure~\ref{fig:FCS++} shows that FCS++ is slightly better, with 38 data sets above the line. And Figure~\ref{fig:TWOFCS} shows that FCS++ achieves a better result than FCS+ on $38$ of the $48$ data sets.
\\\indent \textbf{Time complexity analysis}: Assume that $n$ and $p$ are the number and the length of the finite time series respectively and $c$ is the number of clusters. All three algorithms use the SBD function to measure dissimilarity between data points and centroids. The SBD function requires $O(plog(p))$ time to calculate this measurement. The implementation of FCM as shown in Algorithm~\ref{alg:A1} is $O(npc^2)$. k-Shape uses the k-means algorithm as its underlying clustering algorithm, where the time complexity of k-means is $O(npc)$. Now, the time complexity for the FCS+ algorithm can be calculated as $O(ncplog(p))$ time. In FCS++ and k-Shape clustering algorithms, there is a refinement step, which for every cluster calculates matrix $M$ with $O(p^2)$ time complexity, and then computes an eigenvalue decomposition on $M$ with $O(p^3)$ time complexity. Therefore, the complexity of the refinement step is $O(max\{np^2,cp^3\})$. As a result, the per iteration time complexity of FCS++ and k-Shape are $O(max\{nc^2plog(p), np^2, cp^3\})$ and $O(max\{ncplog(p), np^2, cp^3\})$ time respectively. Thus, all three algorithms are linear in the number of time series, and the major portion of the computational cost rests with $p$, the length of the time series.  
\\\indent Table~\ref{table:tabel2} shows the average CPU time taken by each of the three algorithms to evaluate the Rand index for 10 runs of FCM on each of the $48$ data sets. FCS+ takes roughly $9.5$ secs per run, whereas the other two required about three times that, FCS++ topping out at about $25$ secs per run. This is easy to understand: FCS+ does not use SE Algorithm~\ref{alg:A3}, so it is considerably less expensive computationally than the other two algorithms. Overall, these algorithms are reasonable fast for the $48$ data sets used in our experiments.
%------------------------------------------------------------------------------------------
%                   Table 2
%-------------------------------------------------------------------------------------------
\begin{table}
\centering
	\caption{Average run time for 480 values of the Rand Index (48 data sets, 10 runs each)}
	\begin{tabular*}{.8\linewidth}{@{\extracolsep{\fill}}l c }
		 \toprule
		Algorithm& Ave. CPU Time (in seconds)\\ \midrule
		FCS+ & 9.51  \\ \midrule
		k-Shape & 22.95 \\ \midrule
		FCS++ & 25.04\\
		 \bottomrule
	\end{tabular*}
	\label{table:tabel2}
\end{table}
%---------------------------------------------------------------------------------------------
%-------------------------------------------------------------------------------------------
%                  TABLE WILCXON TEST
%---------------------------------------------------------------------------------------
\begin{table*}[t]
  \caption{Wilcoxon test to compare three clustering methods in terms of accuracy based on $RI$, $ARI$, and $NMI_{M}$ validity method. $R^{+}$ corresponds to the sum of the ranks for
the method on the left and $R^{-}$ for the right}
\begin{tabular*}{\linewidth}{@{\extracolsep{\fill}}lllllllllr}
\toprule
 &\multicolumn{3}{c}{$RI$ Accuracy}  &\multicolumn{3}{c}{$ARI$ Accuracy}   &\multicolumn{3}{c}{$NMI_{M}$ Accuracy}  \\
       \cmidrule(r){2-4}\cmidrule(r){5-7}\cmidrule(r){8-10}
Method    & $R^{+}$& $R^{-}$ & $\rho$-value & $R^{+}$& $R^{-}$ & $\rho$-value & $R^{+}$& $R^{-}$ & $\rho$-value\\
\midrule
k-Shape vs FCS++    & 215    & \textbf{913} & $1.304e^{-04}$  & 66.5    & \textbf{1061.5} & $1.403e^{-07}$ &75  & \textbf{1053} & $1.426e^{-07}$\\
k-Shape vs FCS+    &238  & \textbf{813}& $3.309e^{-04}$ &154 & \textbf{897}& $6.848e^{-06}$ & 81  &\textbf{970}& $1.991e^{-07}$  \\
FCS++ vs FCS+     & \textbf{931} & 245& $2.608e^{-04}$ & \textbf{1036.5} & 139.5& $4.199e^{-06}$ & \textbf{1009}& 167& $4.199e^{-05}$ \\
 \bottomrule
\end{tabular*}
\label{table:wil}
\end{table*}
%-------------------------------------------------------------------------------------------
\\\indent\textbf{Statistical analysis:} We used two statistical tests that assess the statistical significance of the performance of the various methods. First, we perform pairwise comparisons between different methods using the Wilcoxon signed-rank test\cite{jadid13}. The test returns a $\rho$-value associated with each comparison, representing the lowest level of significance of a hypothesis that results in a rejection. This value can be used to determine whether two algorithms have significantly different performance and to what extent. For all the comparisons in this study the significance level $\alpha$ is set to $0.05$. Table~\ref{table:wil} summarizes these comparisons on all the accuracy results from the three algorithms over 48 datasets. In the three mentioned tables, the $\rho$-value is less than the significance level ($\alpha=0.05$), which means that there is a significant difference between the accuracy of algorithms. To illustrate, consider the \textit{RI Accuracy}, where the first row compares k-Shape with FCS++. The sum of ranks for k-Shape is less than the sum of ranks for FCS++. Therefore, the accuracy of FCS++ is better than k-Shape clustering.
%-------------------------------------------------------------------------------------------
\indent Another statistical test that enables us to compare the three clustering algorithms is Friedman's test \cite{jadid14}. The Friedman test is a non-parametric statistical test for differences between groups. Figure~\ref{fig:fridman} shows the results of applying Friedman's test to the three clustering algorithms with respect to the three max optimal CVIs used in our study. Friedman's test supports the conclusions drawn from Wilcoxon's test, viz., that FCS++ is superior to FCS+, which is in turn superior to k-Shape.
%--------------------------------------------------------------------------------------
%----------------------------------------------------------------------------------------------
%                   Friedman figure
%---------------------------------------------------------------------------------------------
%----------------------------------------------------------------------------------------------
%----------------------------------------------------------------------------------------------
%               sECTION dISCUSSION
%--------------------------------------------------------------------------------------------
\section{Discussion}\label{sec:discuss}
\begin{figure*}[h]%
	\centering
	\begin{subfigure}[t]{.32\textwidth}
		\includegraphics[width=\textwidth]{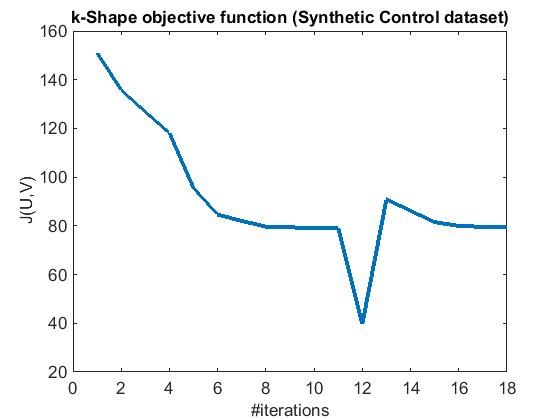}%
		\caption{$J(U,V;X)$ for one run of k-Shape: X=Synthetic Control Data }%
		\label{fig:objcshape}%
	\end{subfigure}\hfill%
	\begin{subfigure}[t]{.32\textwidth}
		\includegraphics[width=\textwidth]{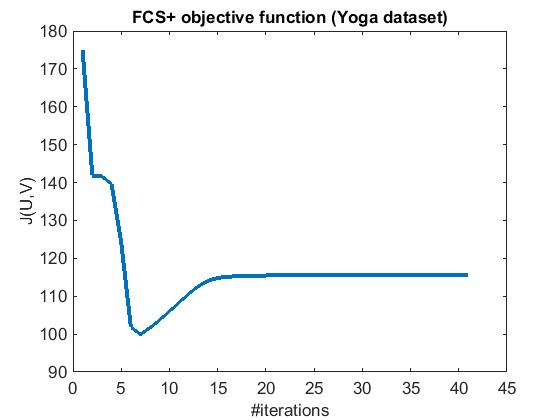}%
		\caption{$J(U,V;X)$ for one run of FCS+: X=Yoga Data}%
		\label{fig:objfcs+}%
	\end{subfigure}\hfill%
	\begin{subfigure}[t]{.32\textwidth}
		\includegraphics[width=\textwidth]{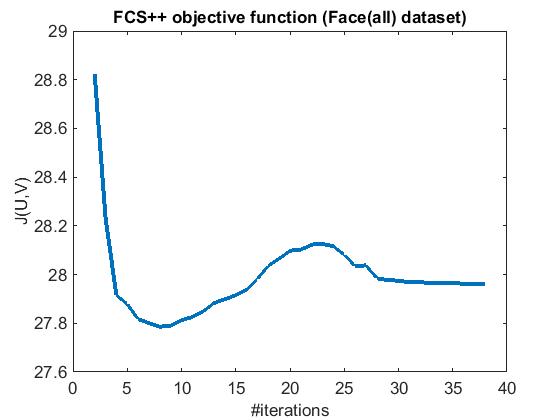}%
		\caption{$J(U,V;X)$ for one run of FCS++: X=Face (All) Data }%
		\label{fig:objfcs++}%
	\end{subfigure}
	\caption{Objective functions for algorithms k-Shape, FCS+,and FCS++, are not always monotone decreasing on successive iterates}
	\label{fig:objective}
\end{figure*}

The authors of \cite{bibi14} assert that k-Shape effectively minimizes the classical HCM objective function $J_{1}(U,P;X)$ when $P$ is computed with SE Algorithm~\ref{alg:A3} and the model norm is replaced by $SBD(x,y)$. However, \textit{(i)} no theory is given to support their statement about minimization; \textit{(ii)} Algorithm~\ref{alg:A2} was not shown to represent an actual metric distance; and \textit{(iii)} it is not clear whether the distances from Algorithm~\ref{alg:A2} should properly be interpreted as squared.
\\\indent The convergence properties of FCM and HCM are well known \cite{bibi33}, and the basic theory for both depends on iterate sequences having the descent property, i.e., that for successive iterates, $J_{m}(U_{t+1},V_{t+1})\leq J_{m}(U_{t},V_{t})$. The objective functions that might be optimized by these three heuristic algorithms are all related in principle to the LSE function at Equation~(\ref{eq:fcm2}), i.e.,
%-------------------------------------------------------
\begin{align}
J_{m}(U,P;X)=\displaystyle\sum_{k=1}^{n}\displaystyle\sum_{i=1}^{c}(u_{ik})^{m}[\Delta(x_{k},p_{i})]^{2}.\nonumber
\end{align}
%-------------------------------------------------------
Table~\ref{table:tabel3} lists the components of $J_{m}(U,P;X)$ for each of the three methods tested in this paper, along with the corresponding information for HCM and FCM. In all cases $P$ is a set of $c$ vectors $V\subset\Re^{cp}$. For FCM, HCM and FCS+ they are unconstrained centroids in $p$ space; for k-Shape and FCS++ the $V$'s are found by shape extraction (Algorithm~\ref{alg:A3}).
\\\indent Do any of the heuristic methods discussed here generate iterative sequences that endow the objective function obtained by replacing $V\subset\Re^{cp}$ and $\Delta(x_{k},p_{i})^{2}$ in Equation~(\ref{eq:fcm2}) with the parameters shown in Table~\ref{table:tabel3} with the descent property? The answer is no. The last column $(\downarrow?)$ of Table~\ref{table:tabel3} summarizes what is known about the descent property for each of the five methods.
\\\indent Figure~\ref{fig:objective} plots presumptive objective functions based on the parameters in Table~\ref{table:tabel2} for one run of k-Shape, FCS+, and FCS++ on three different data sets. Each view in this figure shows that the objective function for each algorithm is not always monotone decreasing on successive iterates. Consequently, the descent property (and in turn, any convergence theory that requires it) is ruled out by counterexample for all three of these heuristic schemes.
%-----------------------------------------------------------------------------------------------
%                              Figure objective
%---------------------------------------------------------------------------------------------
 %-----------------------------------------------------------------------------------------
 %                                    Table 3
 %-----------------------------------------------------------------------------------------
 \begin{table}[ht]
 	\caption{Presumptive Objective Function Components}% title of Tab
 	\begin{tabular*}{\linewidth}{@{\extracolsep{\fill}}c c c c c}% centered columns (4 columns)
 		 \toprule                  %inserts double horizontal lines
 		Method & U & $[\Delta(x_{k},p_{i})]^{2}$ & P & $\downarrow?$ \\[0.5ex]% inserts table
 		%heading
 		 \midrule% inserts single horizontal line
 		$FCM(m>1)$& $M_{fcn}$ & $\|x-v\|_{A}^{2}$ & V& \cmark \\[2ex]% inserting body of the table
 		$FCM(m=1)$ &$M_{hcn}$ & $\|x-v\|_{A}^{2}$ & V& \cmark  \\[0.2ex]
 		$(c-means)$&          &                   &   &      \\[1ex]
 		$k-Shape$ & $M_{hcn}$& $SBD(x_{k},p_{i})$ &SE V& \xmark \\[1ex]
 		$FCS+$ & $H_{mm}(U)$ & $SBD(x_{k},p_{i})$  & V &  \xmark\\[1ex]% [1ex] adds vertical space
 		 %inserts single line
 		$FCS++$ & $H_{mm}(U)$ & $SBD(x_{k},p_{i})$  & SE V &  \xmark\\[1ex]% [1ex] adds vertical space
 		 \bottomrule
 	\end{tabular*}
 	\label{table:tabel3} % is usedto refer this table in the text
 \end{table}
 %----------------------------------------------------------------------------------------------
%----------------------------------------------------------------------------------------------
%                    CONCOLUSION
%---------------------------------------------------------------------------------------------
\section{CONCLUSIONS AND FUTURE REASERCH}\label{sec:conclude}
Two new fuzzy variants of k-Shape were defined. FCS+ arises by substituting $SBD(x_{k},p_{i})$ for $\parallel x-v\parallel_{A}^{2}$ in FCM. FCS++ is FCS+ with the additional substitution of SE V for the prototypes in FCM. For FSC++, partitioning is adapted to the k-Shape algorithm by hardening the fuzzy partitions produced by FCM at each assignment step.
We performed the same experiments with $48$ labeled waveform data sets that were done using k-Shape in \cite{bibi14}, and compared the two fuzzy methods to k-Shape using four crisp external cluster validity indices and two statistical tests of significance. All four indices indicate that FCS++ performs somewhat better than FCS+, and in turn, FCS+ is slightly superior to k-Shape.
While our numerical results are encouraging, they are by no means definitive. More experiments are needed with other waveform data, including unlabeled data. Another avenue of pursuit is the theory: is SBD a metric? Is there any convergence theory for k-Shape, FCS+ or FCS++? Figure~\ref{fig:objective} suggests that the short answer is "probably not." Moreover, any satisfactory convergence theory has to overcome the fact that the SBD and SE schemes are computer programs, not functions. Finally, we have not capitalized fully on the fuzzy information available at each assignment step of the two new methods, instead opting to harden $U$ so that $H_{mm}(U)$ can be used directly by Algorithm~\ref{alg:A3}. Surely there is a better use of this information? This is the enterprise we will turn to next.
\ifCLASSOPTIONcaptionsoff
  \newpage
\fi
%\bibliographystyle{IEEEtran}
\bibliographystyle{IEEEtran}
\bibliography{IEEEabrv}
%\end{IEEEbiography}
%\newpage
%\vfill
\end{document}